\documentclass[preprint,11pt]{elsarticle}

\usepackage[utf8]{inputenc} 
\usepackage[linesnumbered,ruled,vlined]{algorithm2e} 
\usepackage{booktabs} 
\usepackage{hyperref} 

\hypersetup{colorlinks,citecolor=black,filecolor=black, %
linkcolor=black,urlcolor=blue}

\usepackage{amssymb}

\journal{Computers \& Geosciences}

\begin{document}

\begin{frontmatter}



\title{An automatic method for segmentation of fission tracks in epidote 
crystal photomicrographs\footnote{\textit{Published in Computers \& Geosciences (August 2014).} The final publication is available at
\url{http://dx.doi.org/10.1016/j.cageo.2014.04.008}.}}


\author[addr1]{Alexandre Fioravante de Siqueira}
\ead{siqueiraaf@gmail.com}

\author[addr1]{Wagner Massayuki Nakasuga}
\ead{wamassa@gmail.com}

\author[addr2]{Aylton Pagamisse}
\ead{aylton@fct.unesp.br}

\author[addr1]{Carlos Alberto Tello Saenz}
\ead{tello@fct.unesp.br}

\author[addr1]{Aldo Eloizo Job\corref{cor}}
\ead{job@fct.unesp.br}

\cortext[cor]{Corresponding author. Phones: +55(18)3229-5776 / +55(18)3229-5775.}

\address[addr1]{DFQB - Departamento de Física, Química e Biologia \\
				FCT - Faculdade de Ciências e Tecnologia \\
				UNESP - Univ Estadual Paulista \\
				Rua Roberto Simonsen, 305, 19060-900 \\
				Presidente Prudente, São Paulo, Brazil}

\address[addr2]{DMC - Departamento de Matemática e Computação \\
				FCT - Faculdade de Ciências e Tecnologia \\
				UNESP - Univ Estadual Paulista \\
				Rua Roberto Simonsen, 305, 19060-900 \\
				Presidente Prudente, São Paulo, Brazil}

\begin{abstract}
Manual identification of fission tracks has practical problems, such as 
variation due to observer-observation efficiency. An automatic processing 
method that could identify fission tracks in a photomicrograph could 
solve this problem and improve the speed of track counting. However, 
separation of nontrivial images is one of the most difficult tasks in 
image processing. Several commercial and free softwares are available, 
but these softwares are meant to be used in specific images. In this 
paper, an automatic method based on starlet wavelets is presented in 
order to separate fission tracks in mineral photomicrographs. Automatization 
is obtained by Matthews correlation coefficient, and results are evaluated 
by precision, recall and accuracy. This technique is an improvement of a 
method aimed at segmentation of scanning electron microscopy images. This 
method is applied in photomicrographs of epidote phenocrystals, in which 
accuracy higher than 89\% was obtained in fission track segmentation, even 
for difficult images. Algorithms corresponding to the proposed method 
are available for download. Using the method presented here, an user 
could easily determine fission tracks in photomicrographs of mineral samples.
\end{abstract}

\begin{keyword}
Epidote \sep Fission Track \sep Image Processing \sep Optical 
Microscopy \sep Wavelets


\end{keyword}

\end{frontmatter}


\section{Introduction} 
\label{INTRODUCTION}

Fission tracks are dislocated zones caused by nuclear fragments released 
in spontaneous fission of uranium-238. Information about fission tracks 
can be related to geologic events, as mineral crystallization age, geologic 
fault zones and thermal events\cite{WAGNER1992}.

Tracks crossing a polished mineral surface can be etched and visualized 
under an optical microscope, and its selection is based on the following 
relatively simple criteria\cite{FLEISCHER1964,PRICE2005}:

\begin{itemize}
	\item Fission tracks form straight line defects of a limited length ($< 20 \mu m$);
	\item They exhibit no preferred orientation and disappear after suitable heating.
\end{itemize}

Manual identification of fission tracks has some practical problems, such as 
variation due to observer-observation efficiency. Also, Gleadow 
\textit{et al.}\cite{GLEADOW2009} list some problems in discrimination of 
fission tracks from non-track defects as polishing scratches and resolving 
multiple track overlaps and small tracks amongst a similarly sized 
background of surface defects.

An automatic identification of fission track could solve this problem and 
improve the speed of track counting. Image processing can be used to 
automatize such task; however, separation of nontrivial images is one 
of the most difficult tasks in image processing\cite{GONZALEZ2008}. Several 
commercial and free softwares are available for this purpose. Nonetheless, 
these softwares are meant to be used in specific images\cite{USAJ2011}.

\subsection{Proposed methodology}
\label{PROPMET}

In this paper we propose an automatic method based on starlet wavelets, 
in order to segment fission tracks in images of natural minerals obtained 
by optical microscopy. Commonly used objective lenses (dry or oil immersion 
type) have total magnification up to 1500 times. In combination with a 
reflected-transmitted light system, it is possible to analyze fission 
tracks\cite{WAGNER1992,GLEADOW1986}.

The proposed approach consists of applying starlet wavelets in a sample 
image to obtain its detail decomposition levels. Based on information 
retrieval (precision, recall and accuracy) and Matthews correlation 
coefficient (MCC), the segmentation level that better represents fission 
tracks of the original image is automatically chosen.

This technique is an improvement of a recent study aimed at segmentation 
of scanning electron microscopy images\cite{SIQUEIRA2014}. An application 
of this method is the separation of fission tracks in images of natural 
minerals, such as volcanic glasses, apatite, zircon, muscovite, epidote, 
among others.

In this paper, the proposed methodology is applied to segment fission 
tracks in photomicrographs of epidote crystals. Results presented in this 
study will be used as a basis to develop an open source software capable 
of extracting fission tracks from images of natural mineral samples in 
order to establish the age of the material using the fission track dating 
method\cite{WAGNER1992}. A prototype of this software, containing the 
algorithms used in this study, is available for download on this journal 
website (see \ref{SUPPLMAT}).

The remainder of this paper follows. Section \ref{MATANDMET} introduces 
the material used in this study and starlet wavelets, as well as an 
overview of evaluation and automatization methods. Next, Section \ref{RESULTS} 
presents the results from this method application in test photomicrographs. 
Moreover, the method performance is discussed. In the following, Section 
\ref{CONCLUSION} presents the final considerations about this study. 
Finally, \ref{SUPPLMAT} explains where to obtain the cited 
algorithms and how to use them.

\section{Material and Methods} 
\label{MATANDMET}

\subsection{Epidote crystals} 

Epidote is a mineral with monoclinic crystal structure and general formula 
$Ca_{2}(Al,Fe)_{3}Si_{3}O_{12}(OH)$\cite{ITO1950}. According to Poli and 
Schmidt\cite{POLI2004}, it is possible to have epidote formation at 
temperatures of $500 \sim 700^{\circ}C$ (pressure range of 0.2 to 0.6 GPa), 
and also at $720 \sim 760 ^{\circ}C$ (pressure range of 1.6 to 3 GPa). Their 
formation is given by different means. One of them is deuteric action, 
during the late phase of magmatic crystallization stage, by regional 
metamorphism and hydrothermal activity, \textit{i.e.} percolation of solutions 
which chemically react with the rock through fractures, often in 
temperatures between $300$ and $500^{\circ}C$\cite{BAR1974}.

In order to evaluate the proposed methods, we used a data set consisting 
of 45 images. These images were obtained from epidote phenocrystals using 
a Carl ZEISS optical microscope with Axiocam Imager.M1m system, nominal 
magnification factor of $1000X$ (dry) and transmitted light.

\subsection{Starlet transform} 

Starlet wavelet transform is an isotropic redundant wavelet based on the 
algorithm ``\textit{à trous}'' (with holes)\cite{HOLSCHNEIDER1990,SHENSA1992}. 
The construction of this wavelet is given by its scale and wavelet functions, 
respectively $\phi_{1D}$ and $\psi_{1D}$ (Eqs. \ref{STARSCALE} and \ref{STARWAVE}, 
\cite{STARCK2006,STARCK2010}), where $\phi_{1D}$ is the third order B-spline 
($B_{3}$-spline).

\begin{eqnarray}
\phi_{1D}(t) = \frac{1}{12}\left(|t-2|^3-4|t-1|^3+6|t|^3-4|t+1|^3+|t+2|^3\right) \label{STARSCALE} \\
\frac{1}{2}\psi_{1D}\left(\frac{t}{2}\right) = \phi(t)-\frac{1}{2}\phi\left(\frac{t}{2}\right) \label{STARWAVE}
\end{eqnarray}

An extension to two dimensions is achieved by a tensor product (Eq. \ref{STAR2D}),
\begin{eqnarray}
\phi(t_1,t_2) = \phi_{1D}(t_1)\phi_{1D}(t_2) \nonumber\\
\frac{1}{4}\psi\left(\frac{t_1}{2},\frac{t_2}{2}\right) = \phi(t_1,t_2)-\frac{1}{4}\phi\left(\frac{t_1}{2},\frac{t_2}{2}\right) \label{STAR2D}
\end{eqnarray}

These wavelets were successfully employed in analysis of astronomical 
\cite{STARCK2006,STARCK2010,STARCK2011} and biological \cite{GENOVESIO2003} 
images, being suitable to evaluate images that contains isotropic objects. 
Isotropic transforms retrieve only one detail set per level instead of 
several detail sets (\textit{e.g.} Daubechies wavelets have horizontal, vertical 
and diagonal detail levels), facilitating the interpretation of results. The 
properties of this wavelet (isotropy, redundancy, translation-invariance) 
make it a good alternative in image processing and pattern recognition.

Similarly to Eq. \ref{STAR2D}, the pair of filters $(h, g)$ related to 
this wavelet is (Eq. \ref{FILTERPAIR}, \cite{STARCK2010})
\begin{eqnarray}
h_{1D}[k] = [\begin{array}{ccccc} 1 & 4 & 6 & 4 & 1\end{array}]/16, k = -2,...,2 \nonumber\label{H1D}\\ 
h[k,l] = h_{1D}[k]h_{1D}[l] \nonumber\\
g[k,l] = \delta[k,l]-h[k,l] \label{FILTERPAIR}
\end{eqnarray}
where $\delta$ is defined as $\delta[0,0] = 1$, $\delta[k,l] = 0$ for $[k,l] \neq 0$. 
From Eqs. \ref{STAR2D} and \ref{FILTERPAIR}, detail wavelet coefficients 
are obtained from the difference between the current and previous resolutions.

Starlet transform application is given by a convolution between an input 
image $c_{0}$ and the finite impulse response (FIR) filter derived from 
$\phi$ (Eq. \ref{H2D} \cite{STARCK2010}),

\begin{center}
\begin{equation}
h = \left[\begin{array}{ccccc} \frac{1}{256} & \frac{1}{64} & \frac{3}{128} & \frac{1}{64} & \frac{1}{256} \\
\frac{1}{64} & \frac{1}{16} & \frac{3}{32} & \frac{1}{16} & \frac{1}{64} \\ 
\frac{3}{128} & \frac{3}{32} & \frac{9}{64} & \frac{3}{32} & \frac{3}{128} \\
\frac{1}{64} & \frac{1}{16} & \frac{3}{32} & \frac{1}{16} & \frac{1}{64} \\ 
\frac{1}{256} & \frac{1}{64} & \frac{3}{128} & \frac{1}{64} & \frac{1}{256} \end{array}\right]
\label{H2D}
\end{equation}
\end{center}

This convolution results in a set of smoothing coefficients which correspond 
to the first decomposition level, $c_{1}$. Detail wavelet coefficients of 
the first decomposition level are obtained from the difference 
$w_{1} = c_{0} - c_{1}$.

Let $L$ be the last resolution level. Therefore, resolution levels can be 
calculated by:
\begin{eqnarray*}
c_{j} = c_{j-1} * h, \\
w_{j} = c_{j-1} - c_{j},
\end{eqnarray*}
\noindent with $j = 0, \ldots, L$, and $*$ the convolution operation. The set 
$W = \{w_{1}, \ldots, w_{L}, c_{L}\}$
obtained by these operations is the starlet transform of the input image.

\subsection{Evaluation of the results} 
\label{SECEVAL}

In order to evaluate the proposed methodology, we employed precision, 
recall and accuracy\cite{OLSON2008,WANG2013}. These values are based on 
the concepts of true positives (TP), true negatives (TN), false positives 
(FP) and false negatives (FN).

Fission tracks in an image sample are represented in a ground truth (GT) 
image: black represents the background, whereas white represents 
fission tracks in this image. Comparing an input image and its ground 
truth, TP, TN, FP and FN values could be established as:

\begin{itemize}
\item TP: pixels correctly labeled as fission tracks.
\item FP: pixels incorrectly labeled as fission tracks.
\item FN: pixels incorrectly labeled as background.
\item TN: pixels correctly labeled as background.
\end{itemize}

Based on these considerations, precision (retrieved pixels that are 
relevant), recall (relevant pixels that were retrieved) and accuracy 
(proportion of true retrieved results) are defined:

\begin{eqnarray}
precision &=& \frac{TP}{TP+FP} \nonumber \\
recall &=& \frac{TP}{TP+FN} \nonumber \\
accuracy &=& \frac{TP+TN}{TP+TN+FP+FN} \nonumber
\end{eqnarray}

\subsubsection{Method automatization} 

In order to establish the optimal level for method application, Matthews 
correlation coefficient (MCC, \cite{MATTHEWS1975}) is used. MCC uses TP, 
TN, FP and FN, and may offer an evaluation of the segmentation correctness:

\begin{center}
\begin{equation}
MCC = \frac{TP*TN-FP*FN}{\sqrt{(TP+FN)(TP+FP)(TN+FP)(TN+FN)}} \label{MCC}
\end{equation}
\end{center}

This coefficient measures how variables tend to have the same sign and 
magnitude, where $1$, zero and $-1$ indicates perfect, random and imperfect 
predictions, respectively\cite{BALDI2000}.

Thereby, automatic retrieval of the optimal segmentation level is achieved by:
\begin{itemize}
\item applying the method for $L$ desired starlet decomposition levels;
\item comparing method segmentation results with the image GT and obtaining 
TP, TN, FP and FN;
\item calculating MCC (Eq. \ref{MCC}) for each $L$.
\end{itemize}
As the optimal segmentation level is reached, values yielded by MCC become 
higher. Then, the best segmentation level obtained by this method is the 
one that returns the highest MCC value.

\subsection{Method overview} 

The proposed automatic segmentation method is defined as follows:
\begin{itemize}
\item Starlet transform is applied in an input image $c_{0}$, resulting 
in $L$ detail levels: $D_{1}, \cdots, D_{L}$, where $L$ is the last 
desired resolution level.
\item First and second detail levels, $D_{1}$ and $D_{2}$, are ignored 
due to the large amount of noise; third to $i$ detail levels are summed 
($Ri = D_{3} + \cdots + D_{i}$), where $3 \leq i \leq L$. This is the 
result of the method application related to starlet level $i$.
\item TP, TN, FP and FN are obtained comparing Ri with its GT (see 
Section \ref{SECEVAL}). MCC (Eq. \ref{MCC}) is calculated using these 
values.
\item Therefore, the optimal segmentation level is the one that has a 
higher MCC between the $R_{L}$ levels of method application.
\end{itemize}

To apply this method, one can use the pseudocodes given in Algorithm 
\ref{ALGORITHM1}. Also, the source code of these algorithms in 
Matlab\footnote{Matlab is a numerical computing environment and a 
programming language developed by MathWorks. A trial version could be 
requested at \url{https://www.mathworks.com}.}/Octave\footnote{GNU Octave is 
an open source high-level interpreted language intended primarily for 
numerical computation. Download available freely at 
\url{http://www.gnu.org/software/octave/download.html}.} programming language 
is available (see \ref{SUPPLMAT}).

\begin{algorithm} 
\SetAlgoNoEnd
\DontPrintSemicolon
\KwIn{\\
$\bullet$ A grayscale image, $c_{0}$.\\
$\bullet$ A ground truth image, $gtc_{0}$.\\
$\bullet$ Number of resolutions to be calculated, $L$.}
\KwOut{\\
$\bullet$ Detail coefficients from starlet transform, $w_{j}$.\\
$\bullet$ An optimal image that presents fission tracks contained in the
original image, $optft$. \\
$\bullet$ Matthews Correlation Coefficient (Eq. \ref{MCC}) between $gtc_{0}$ and
the algorithm result for each level $imgft_i$, $MCC_{i}$, 
with $i = 3,...,L$.}
\textbf{mirroring(}$c_{0}$\textbf{);}
\For{$j \gets 1$ \textbf{to} $L$} {
	$h \gets$ \textbf{hgen(}j\textbf{);} \\
	$c_{j} \gets$ \textbf{convolution(}$c_{j-1},h$\textbf{);} \\
	$w_{j} \gets c_{j-1}-c_{j}$\textbf{;} \\
	\textbf{unmirroring(}$c_{j}$\textbf{);} \\
	\textbf{increment(}$j$\textbf{);} \\
}
\textbf{initialize} $sum$ \textbf{to} 0\textbf{;}

\For{$i \gets 3$ \textbf{to} $L$} {
	\For{$j \gets 3$ \textbf{to} $i$} {
		$sum \gets sum + w_{j};$ \\
		$imgft_{i} \gets sum;$
	}
	$MCC_{i}(imgft_{i}, gtc_{0})$; \\
}
$optft \gets \max(MCC_{i})$; \\
\Return{$w_{j},optft,MCC_{i}$}\;
\caption{Pseudocode for automatic determination of fission tracks in an 
image, based on starlet algorithm application (adapted from \cite{SIQUEIRA2014,STARCK2011}).}
\label{ALGORITHM1}
\end{algorithm}

Starlet transform of $c_{0}$ is given by $W = \{w_{1}, \cdots, w_{L}, c_{L}\}$.
\textbf{hgen()} (Algorithm \ref{ALGORITHM2}), referenced on Algorithm 
\ref{ALGORITHM1}, is applied when $j$ is incremented. For $j>1$, $h$ has $2^{j-1}$ 
zeros between its elements, characterizing the \textit{à trous} transform.

\begin{algorithm} 
\SetAlgoNoEnd
\DontPrintSemicolon
\KwIn{\\
$\bullet$ $h_{1D}$ filter, given by Eq. \ref{H1D}.\\
$\bullet$ Current resolution level, $j$.}
\KwOut{\\
$\bullet$ Filter $h_{2D}$, h.}
\If{$j = 0$}{
	$h \gets h_{1D}$;\;
}
\Else{
	$M \gets$ \textbf{size(}$h_{1D},2$\textbf{)};
	\textbf{initialize} $k$ \textbf{to} $0$;\;
	\For{$i \gets 1$ \textbf{step} $2^{i-1}$ \textbf{to} $M+2^{i-1}*(M-1)$}{
		\textbf{increment(}$k$\textbf{)};\;
		$h(i) \gets h_{1D}(k)$;\;
	}
}
\textbf{initialize} $aux$ \textbf{to} 0;\;
$aux \gets sum(sum(h'*h))$;\;
$h \gets (h'*h)/aux$;\;
\Return{$h$}\;
\caption{\textbf{hgen:} $h$ filter generation and zero-inserting\cite{SIQUEIRA2014}.}
\label{ALGORITHM2}
\end{algorithm}

\section{Experimental results} 
\label{RESULTS}

\begin{figure*}[h] 
\centering
\includegraphics[scale=0.4]{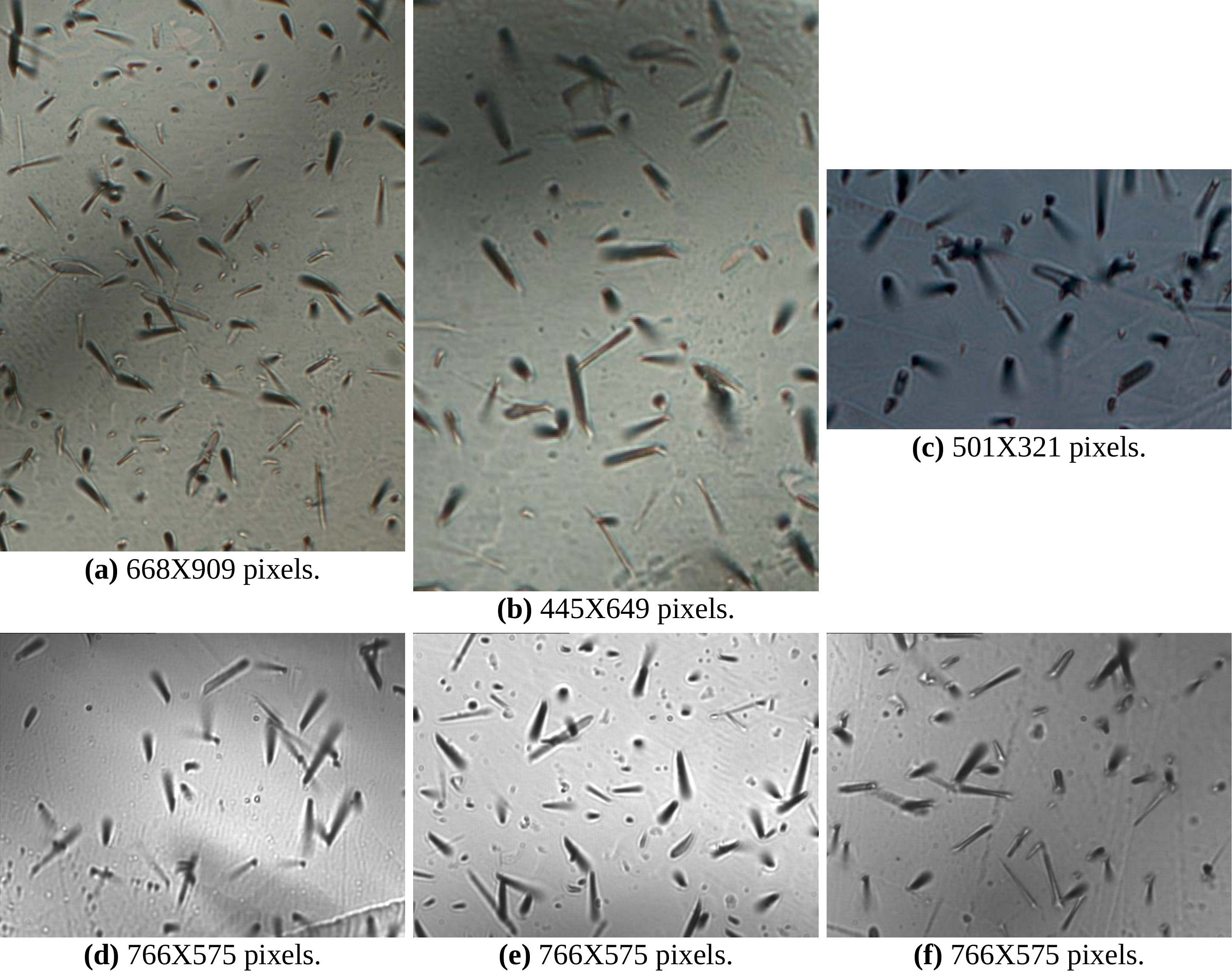}
\caption{Photomicrographs of epidote phenocrystals. Fission tracks are 
shown in sample surface as dark segments. Nominal magnification factor: 
1000X (dry).}
\label{FIGSAMPLES}
\end{figure*}

In order to present the proposed method results, six dataset images with 
different size and fission track distribution are shown (Fig. 
\ref{FIGSAMPLES}). The darker regions of these images correspond to 
fission tracks in epidote surface.

\begin{figure*}[h]
\centering
\includegraphics[scale=0.45]{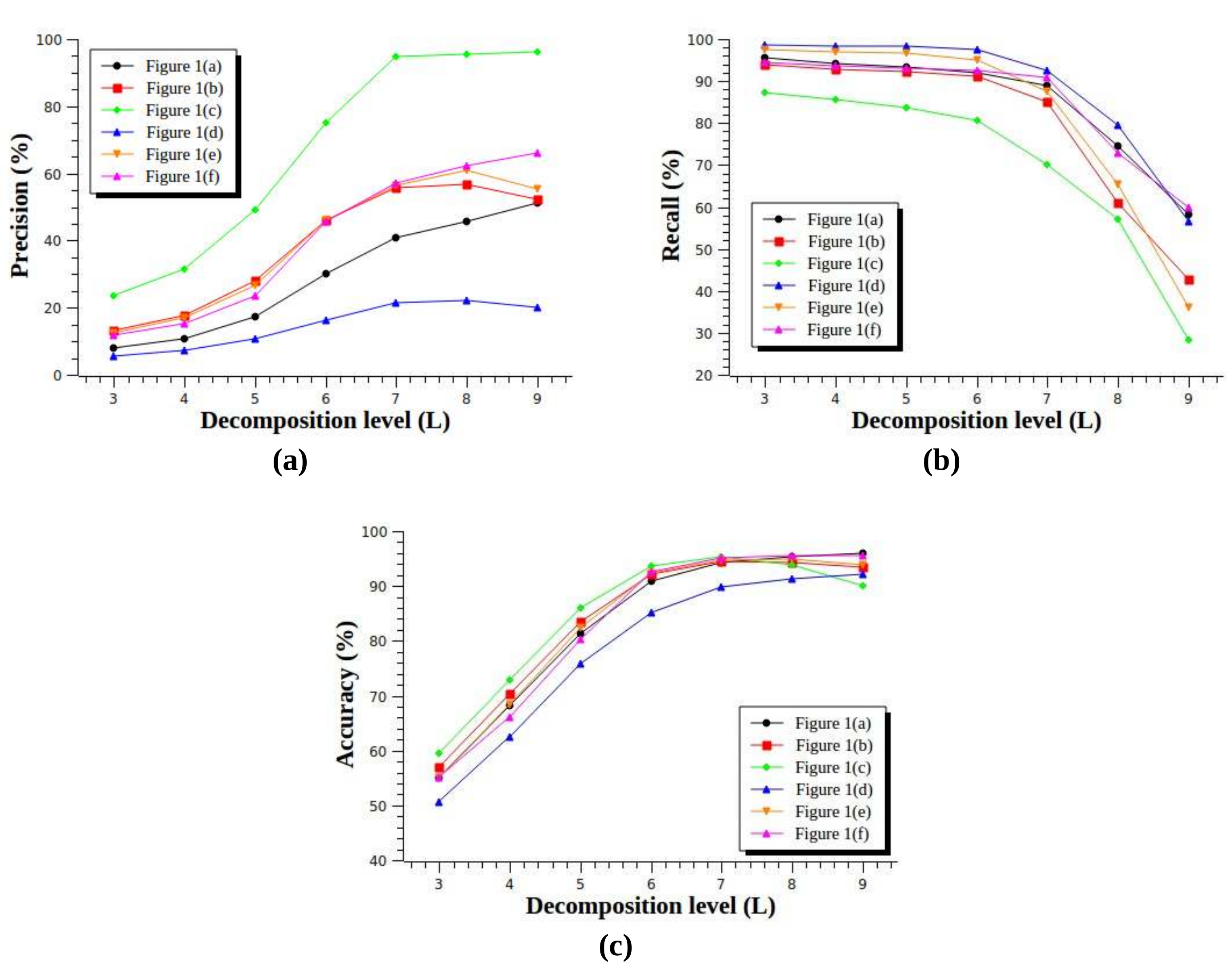}
\caption{Precision, recall and acuracy values obtained from Fig. \ref{FIGSAMPLES}.}
\label{FIGPRECRECACC}
\end{figure*}

The proposed method was applied in the test images with $L = 3$ to $L = 9$. 
The optimal segmentation level was obtained from MCC, for each image. 
Also, precision, recall and accuracy were obtained for each level (Fig. 
\ref{FIGPRECRECACC}), in order to evaluate the method performance. For a 
satisfactory segmentation degree, an optimal ratio between precision and 
recall becomes necessary.

One could see from Fig. \ref{FIGPRECRECACC} that accuracy and precision 
increases until level $L = 7$. On the other hand, recall decreases as 
$L$ increases.

\begin{figure*}[htb]
\centering
\includegraphics[scale=0.4]{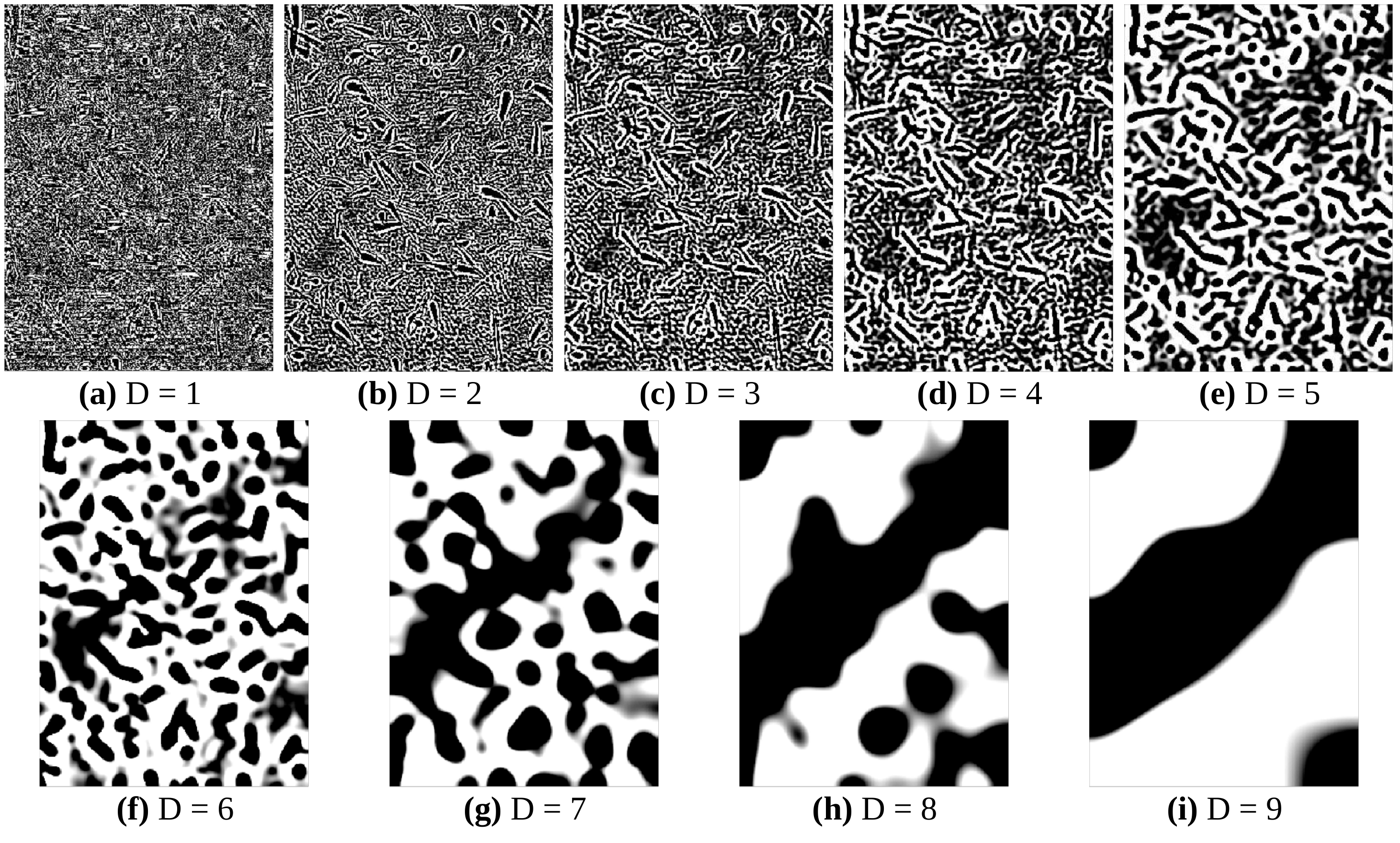}
\caption{Starlet detail decomposition levels of Fig. \ref{FIGSAMPLES}(a). 
$D = 1$ and $D = 2$ were disregarded in method application due to noise 
amount. Although higher detail levels tend to aggregate, reducing 
segmentation accuracy, these levels present better information about 
fission tracks.}
\label{FIGSTARLET}
\end{figure*}

Fig. \ref{FIGSAMPLES}(a) will be used to introduce the proposed method. 
According to the first step, starlet transform detail levels are obtained 
from the input image (Fig. \ref{FIGSTARLET}).

\begin{figure*}[htb]
\centering
\includegraphics[scale=0.4]{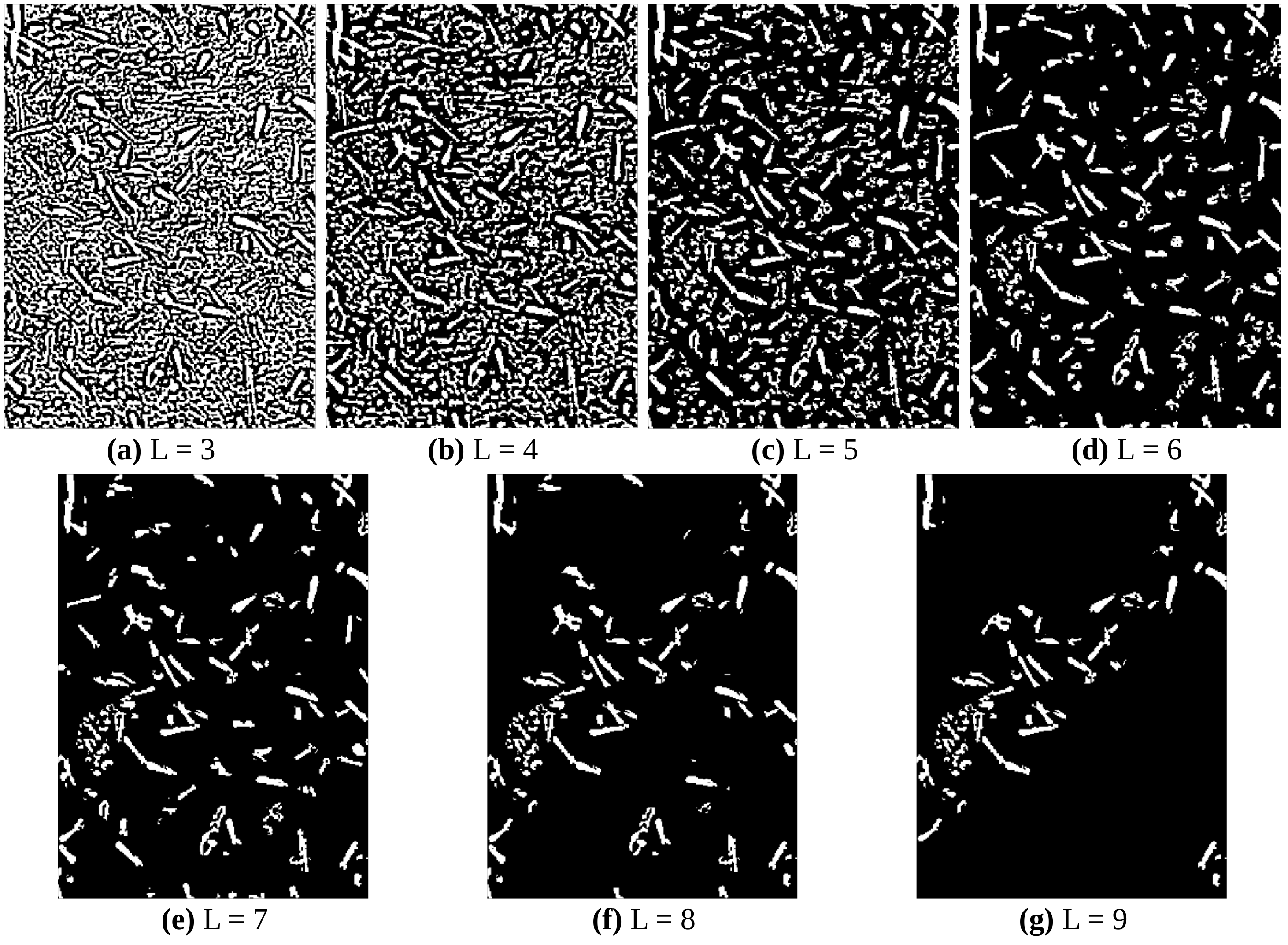}
\caption{Segmentation output of the proposed method. Different levels, 
from $L = 3$ to $L = 9$ were considered.}
\label{FIGOUTPUT}
\end{figure*}

After starlet application, Algorithm \ref{ALGORITHM1} is applied seven 
times, from $L = 3$ to $L = 9$. For example, method application for 
$L = 6$ consists of:
\begin{itemize}
	\item disregard $D = 1$ (Fig. \ref{FIGSTARLET}(a)) and D = 2 (Fig. \ref{FIGSTARLET}(b)).
	\item sum $D = 3$ (Fig. \ref{FIGSTARLET}(c)) to $D = 6$ (Fig. \ref{FIGSTARLET}(f)): $\sum D_{i}, i = 3,\cdots, 6$.
\end{itemize}
Results of the proposed method are shown as binary images, where fission 
tracks are represented by the white color and background by the black color 
(Fig. \ref{FIGOUTPUT}).

The next step is to determine the optimal segmentation level using MCC 
(Table \ref{TABLE1}). Values presented here are given in 
percentages in order to ease results comprehension. The optimal 
segmentation level for images of Fig. \ref{FIGSAMPLES} is $L = 7$, according 
to MCC; then results obtained with $L = 7$ using the proposed method 
were compared to ground truth (GT).

\begin{figure*}[htb]
\centering
\includegraphics[scale=0.5]{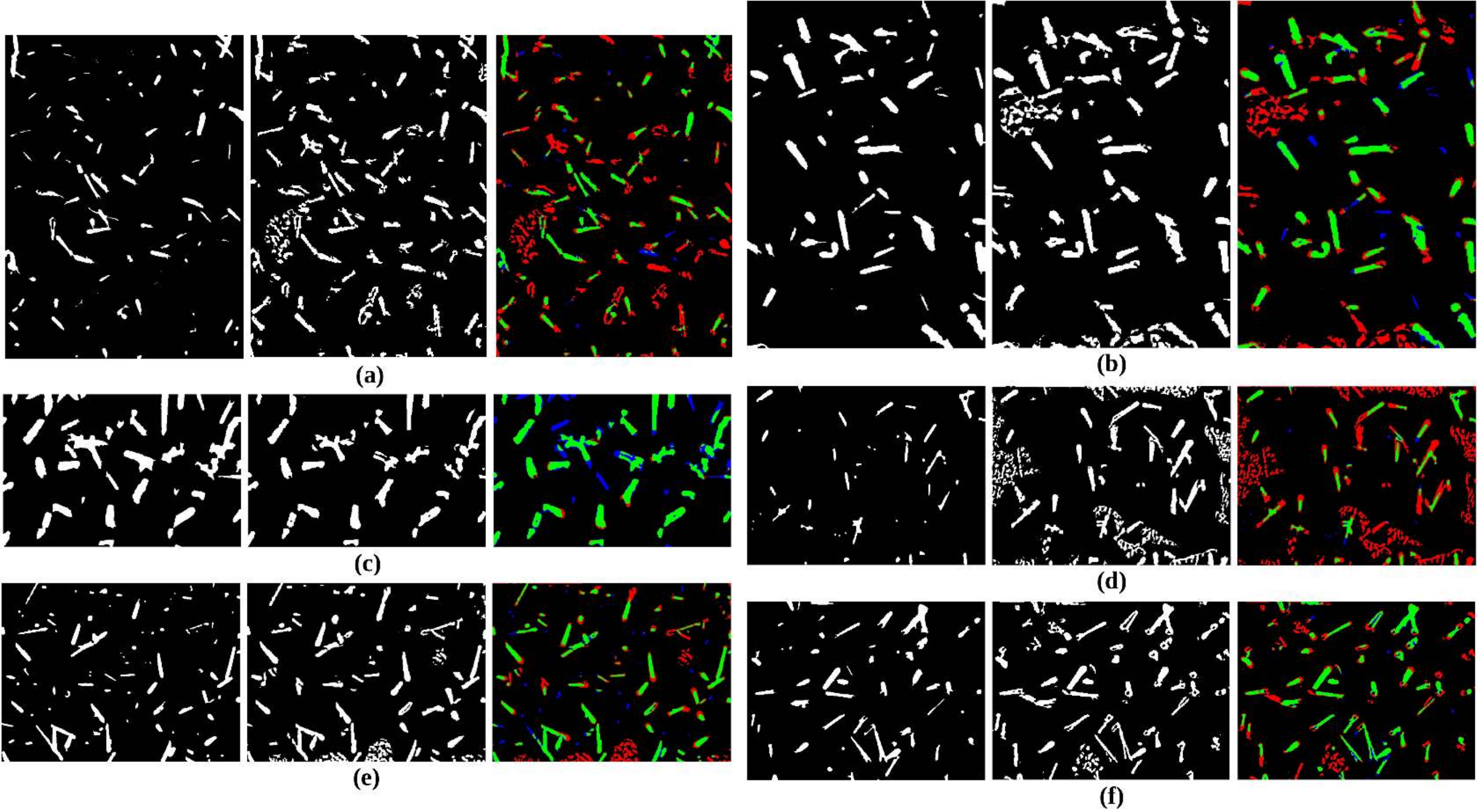}
\caption{First column: Ground truths for Fig. \ref{FIGSAMPLES} images. 
Second column: proposed method output for the optimal decomposition level. 
Third column: a comparison between GT and method output for optimal level 
(Green: TP pixels; blue: FN pixels; red: FP pixels).}
\label{FIGCOMPARISON}
\end{figure*}

GT images obtained from Fig. \ref{FIGSAMPLES} are used to evaluate the 
method performance. These images were obtained manually by a specialist 
using GIMP\footnote{Available freely at http://www.gimp.org/downloads/.}, 
an open source graphics software. TP pixels are shown as green, FN 
pixels as blue and FP pixels as red, to facilitate visualization of 
differences between the images (Fig. \ref{FIGCOMPARISON}).

\begin{table*}[htb]
\caption{MCC obtained from method application for levels $L = 3$ to 
$L = 9$. $L = 7$ was the optimal segmentation level for Fig. \ref{FIGSAMPLES} 
images, according to MCC values.}
{\small
\begin{center}
\begin{tabular}{lccccccc}
\toprule
\bf{MCC (\%)} & \bf{L = 3} & \bf{L = 4} & \bf{L = 5} & \bf{L = 6} & \bf{L = 7} & \bf{L = 8} & \bf{L = 9}\\
\midrule
\bf{Fig. 1 (a)} & 19.32996 & 25.30251 & 35.37489 & 49.55525 & 57.85511 & 56.06013 & 52.62559\\
\bf{Fig. 1 (b)} & 24.22276 & 32.30995 & 45.43835 & 61.52304 & 66.14956 & 55.69191 & 43.78881\\
\bf{Fig. 1 (c)} & 29.07748 & 40.25502 & 56.84173 & 74.15157 & 79.22872 & 71.01649 & 49.15956\\
\bf{Fig. 1 (d)} & 16.09310 & 20.57131 & 27.80563 & 36.43833 & 41.70653 & 39.22664 & 30.58268\\
\bf{Fig. 1 (e)} & 24.60903 & 32.38622 & 45.38681 & 63.17591 & 67.75544 & 60.42723 & 41.72874\\
\bf{Fig. 1 (f)} & 22.82558 & 28.93141 & 40.72572 & 62.05960 & 69.84571 & 64.87040 & 60.59735\\
\bottomrule
\end{tabular}
\end{center}
}
\label{TABLE1}
\end{table*}

Most fission tracks presented in GT images were located using level 
$L = 7$. While precision and recall values vary, accuracy is higher than 
$89\%$ for Fig. \ref{FIGSAMPLES} images (Table \ref{TABLE2}). Comparing 
these results with near levels, $L = 6$ and $L = 8$, one can see that 
accuracy method for Fig. \ref{FIGSAMPLES} using $L = 7$ is higher than 
with $L = 6$, but lower than $L = 8$ for Fig. \ref{FIGSAMPLES}(a),(d),(e) 
and (f). However, the difference is small for these cases, less than 
$1.5\%$. Furthermore, level 8 recall values are smaller than recall for 
$L = 7$ (difference between 14 and $22.5\%$).

\begin{table*}[htb]
\caption{Precision, recall and accuracy values for starlet application 
levels 6, 7 and 8.}
{\small
\begin{center}
\begin{tabular}{lccccccccc}
\toprule
& \multicolumn{3}{c}{\bf{Precision (\%)}} & \multicolumn{3}{c}{\bf{Recall (\%)}} & \multicolumn{3}{c}{\bf{Accuracy(\%)}}\\
\midrule
& \bf{L = 6} & \bf{L = 7} & \bf{L = 8} & \bf{L = 6} & \bf{L = 7} & \bf{L = 8} & \bf{L = 6} & \bf{L = 7} & \bf{L = 8}\\
\midrule
\bf{Fig. 1 (a)} & 30.047 & 40.674 & 45.556 & 91.918 & 88.860 & 74.529 & 90.899 & 94.232 & 95.306 \\
\bf{Fig. 1 (b)} & 46.044 & 55.792 & 56.656 & 91.219 & 84.953 & 61.034 & 92.163 & 94.420 & 94.197 \\
\bf{Fig. 1 (c)} & 74.977 & 94.820 & 95.545 & 80.685 & 70.104 & 56.958 & 93.731 & 95.428 & 93.805 \\
\bf{Fig. 1 (d)} & 16.186 & 21.400 & 22.303 & 97.641 & 92.448 & 79.438 & 85.188 & 89.878 & 91.330 \\
\bf{Fig. 1 (e)} & 46.118 & 56.291 & 60.873 & 95.075 & 87.678 & 65.457 & 92.369 & 94.711 & 94.960 \\
\bf{Fig. 1 (f)} & 45.815 & 57.202 & 62.156 & 92.491 & 90.864 & 72.802 & 92.581 & 95.105 & 95.461 \\
\bottomrule
\end{tabular}
\end{center}
}
\label{TABLE2}
\end{table*}

Accurate results were obtained when images contained better visual state; 
for example, images without grain fractures (as seen in Fig. 
\ref{FIGSAMPLES}(d)). Also, textures in the background may lead to incorrect 
segmentation, thus lowering method accuracy. The red agglomerated regions 
in Fig. \ref{FIGOUTPUT} exhibit this issue.

\section{Conclusion} 
\label{CONCLUSION}

In this study we present a fission track automatic segmentation method 
for photomicrographs, based on starlet wavelets. This method uses starlet 
decomposition detail levels to determine edges of objects in an input 
image. Levels corresponding to noise are discarded and remain levels 
are summed. The method presented here can help the user to determine fission 
tracks in photomicrographs of mineral samples.

Automation is achieved using precision, recall and accuracy, together with 
Matthews correlation coefficient. MCC has proved to be a satisfactory measure 
to the method automation, representing a good balance between precision, 
recall and accuracy.

An application of this method is given here, in epidote crystal images 
obtained by optical microscopy. In this application, the proposed method 
presents a high accuracy degree, even for challenging images.

Algorithms used in this study are available for download. In future studies, 
this methodology will be used in images obtained by different materials, 
in order to estimate related features. Also, from this algorithms, an 
open source software aimed to analyze fission tracks in microscopical 
images will be built.

\section*{Acknowledgements} 
\label{ACKNOWLEDGEMENTS}
 
The authors would like to acknowledge the Brazilian foundations of research 
assistance CNPq, CAPES and FAPESP. This research is supported by FAPESP 
(Procs 2010/20496-2 and 2011/09438-3).

\appendix

\section{Supplementary material}
\label{SUPPLMAT}

To use the available supplementary algorithms, it is necessary to have 
two images: a sample image and its ground truth. These files could be put 
in the same folder of the algorithm files. In Matlab/Octave 
prompt, navigate to the folder that contains the algorithms. Then, type 
the following commands:

\noindent \texttt{> IMG = imread('your\_test\_image');} \\
\texttt{> IMGGT = imread('your\_ground\_truth\_image');} \\
\texttt{> [D,L,COMP,MCC] = main(IMG,IMGGT);} \\
\noindent where \texttt{>} represents the Matlab/Octave prompt.

The software asks the desired application level, and returns starlet 
detail coefficients (\texttt{D}), the method output related to each starlet level 
(\texttt{R}), colored comparison between \texttt{IMG} and \texttt{IMGGT} 
for each method level (\texttt{COMP}) and Matthews correlation coefficients also 
for each level (\texttt{MCC}). These files can be downloaded on this journal 
website.





\bibliographystyle{model1-num-names}
\bibliography{references}


\end{document}